%% file: llm_maprepair_acl.tex
\documentclass[11pt]{article} 
\usepackage{acl}

\usepackage{times}
\usepackage{latexsym}

\usepackage[T1]{fontenc}
\usepackage[utf8]{inputenc}

\usepackage{microtype}

\input{math_commands.tex}

\usepackage{graphicx}
\usepackage{algorithm}
\usepackage{algorithmic}
\usepackage{amsmath}
\usepackage{amsfonts}
\usepackage{amssymb}
\usepackage{booktabs}
\usepackage{multirow}
\usepackage{url}
\usepackage{hyperref}
\usepackage{float}
\usepackage{subcaption}
\usepackage{tabularx}
\newcolumntype{Y}{>{\centering\arraybackslash}X}
\usepackage{xcolor}
\usepackage{natbib}

\title{Constructing coherent spatial memory in LLM agents through graph rectification\thanks{Code and experimental data: \url{https://github.com/nitpicker55555/MapRepair}}}

\usepackage{listings}
\author{
Puzhen Zhang \and Xuyang Chen \and Yu Feng \and Yuhan Jiang \and Liqiu Meng \\
Chair of Cartography and Visual Analytics \\
Technical University of Munich \\
\texttt{\{puzhen.zhang, xuyang.chen, y.feng, yuhan.jiang, liqiu.meng\}@tum.de}
}

\begin{document}
\maketitle

\begin{figure*}[h]
\begin{center}
    \centering
    \includegraphics[width=\linewidth]{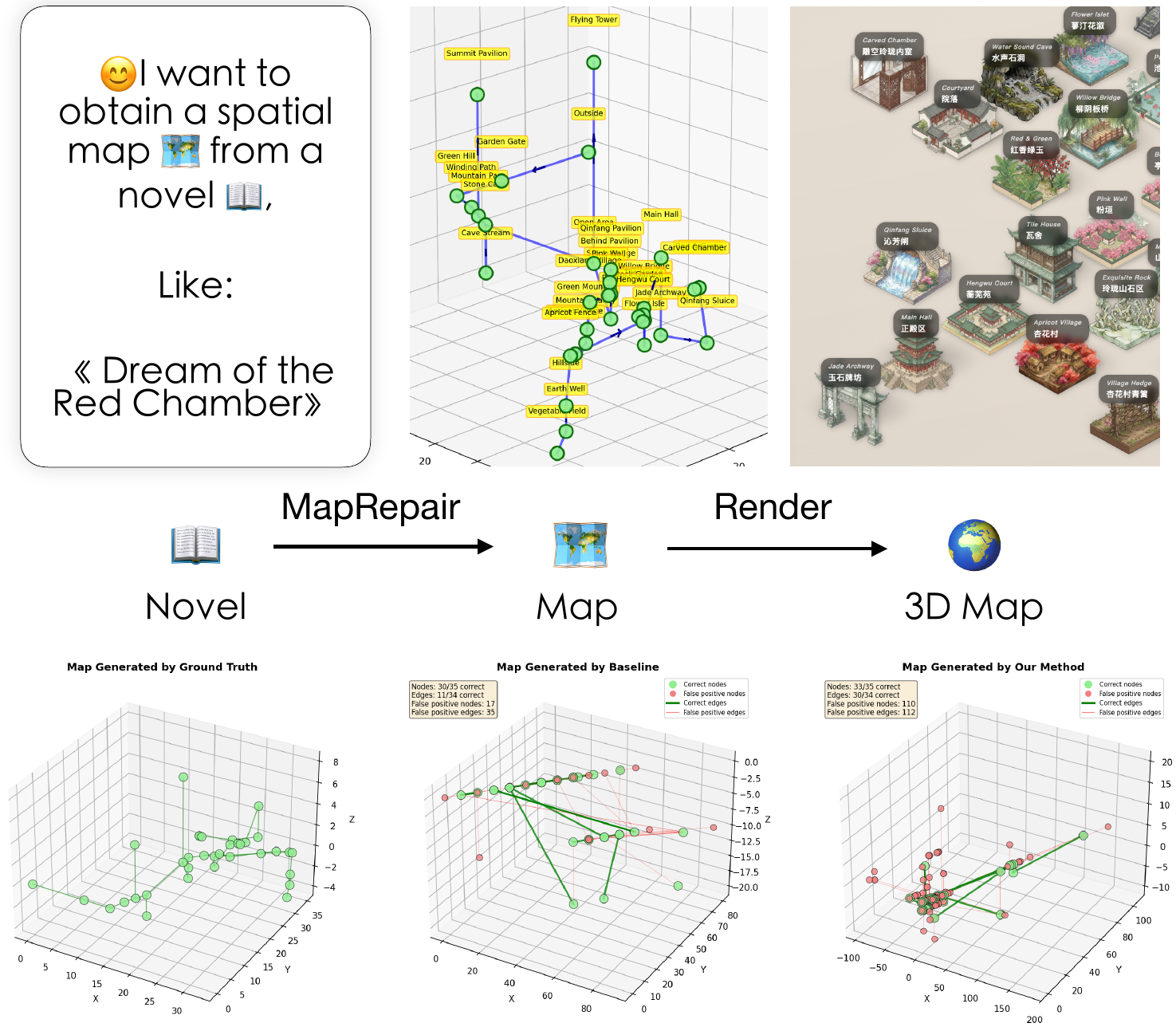}
    \caption{Our framework, LLM-MapRepair, can extract spatial relationships between entities from long-form text and produce structured maps that can subsequently be rendered into high-quality visualizations.}
    \label{fig:problem-overview}
\end{center}
\end{figure*}

\begin{abstract}

Given a map description through global traversal navigation instructions, an LLM can often infer the implicit spatial layout and answer user queries by providing shortest paths. However, such context-dependent querying becomes incapable as environments grow larger, motivating the need for incremental map construction that builds a complete topological graph from stepwise observations. We propose \textbf{LLM-MapRepair}, a framework for LLM-driven construction and map repair, designed to detect, localize, and correct structural inconsistencies in incrementally constructed navigation graphs. Our contributions include a Version Control mechanism for graph construction, an Edge Impact Score for repair prioritization, and a cleaned variant of the MANGO benchmark tailored for LLM-driven map construction and repair. We evaluate the framework on four evaluation settings: a synthetic per-component ablation (gpt-4.1, $n{=}20$ seeds per cell), a cross-vendor sweep over seven LLMs from OpenAI, Anthropic, and Google on both synthetic and TextWorld procedurally-generated text-adventure games, a repair-stage evaluation on all 42 cleaned-MANGO games with non-zero residual conflicts (534 conflicts; three vendors $\times$ three modes plus two non-LLM references), and an end-to-end natural-text deployment on Chapters~16--17 of \emph{Dream of the Red Chamber}. On the DRC deployment, LLM-MapRepair achieves $94.3\%$ node recall ($+8.6$~pp over direct LLM mapping) and $88.2\%$ edge recall ($+55.8$~pp), using GPT-4.1; the recall improvements come with predicted node and edge counts that are roughly $4\times$ the ground-truth counts (Table~\ref{tab:drc_recall}), reflecting the discretization-driven over-generation trade-off we discuss in the Limitations.

\end{abstract}

\section{Introduction}

\label{introduction}
LLMs have shown strong abilities in open-domain reasoning, sequential planning, and text-based navigation. However, in text-processing environments, spatial cognition with LLMs still primarily depends on direct reasoning within the context window \citep{ding2024mangobenchmarkevaluatingmapping}. This approach presents several potential challenges: it may exceed context capacity limitations when processing extensive texts, encounter context forgetting issues when addressing complex problems, and introduce inconsistencies in iterative reasoning processes. Consequently, when tackling complex large-scale spatial problems, adopting a human-like cognitive approach—progressively assembling local spatial cognition to achieve understanding of complex spaces \citep{xia2025llmslearnmapworld}—may constitute a superior solution. For LLMs, this methodology alleviates contextual pressure by incrementally storing local spatial cognition in graph structures, requiring the context to process only current local information. Furthermore, the structured storage through graph representation ensures consistency in search structures and provides error correction capabilities when cognitive biases occur. As demonstrated in Figure \ref{fig:problem-overview}, our LLM-MapRepair framework substantially improves recall, achieving $94.3\%$ node recall and $88.2\%$ edge recall (with the predicted-node and predicted-edge totals reaching roughly $4\times$ the ground-truth counts; see Table~\ref{tab:drc_recall} and Limitations).
Despite this potential, maintaining complex spatial layouts in graph structures remains challenging. Small errors made early can silently propagate, manifesting as conflicts only when sufficient context accumulates. This temporal gap between error introduction and detection is exacerbated by \textit{coupled dependencies}, where one error triggers cascading mistakes. Since most LLMs lack persistent memory or version control, they cannot trace error provenance or reason about when faulty edges were introduced.

To address this challenge, we propose \textbf{LLM-MapRepair}, a modular framework for detecting and repairing topological inconsistencies in navigation graphs constructed by LLM agents. At the core of our method is the \textbf{Version Control}, a versioned graph history that records every modification to the graph, along with its originating observation and time indexed head. Version Control enables time-aware tracing, rollback, and structural comparison, allowing the system to pinpoint the specific actions that introduced inconsistencies, even if they occurred many steps earlier.

To improve the efficiency of graph repair, we introduce an \textbf{Edge Impact Score} that estimates the potential downstream effects of each edge based on reachability, usage frequency, and conflict propagation. High-scoring edges are the most plausible root causes of cascading errors, so the system inspects and rewrites them first; this descending-order inspection strategy is what we call ``Edge-Impact Ranking'' in the experiments.

We evaluate our approach across four evaluation settings (\S\ref{sec:datasets}): controlled-error synthetic graphs for per-component ablation, synthetic and TextWorld-procedurally-generated environments for cross-vendor generalization, cleaned-MANGO LLM-built IF maps for repair-stage evaluation, and Chapters~16--17 of \emph{Dream of the Red Chamber} for natural-text deployment. Experiments show that our method substantially improves structural integrity and overall task performance, especially in cases involving long-range error propagation.
Our contributions are as follows:

\begin{itemize}
    \item We identify a critical limitation of LLM-based agents in long-horizon exploration: their inability to detect and correct accumulated structural errors that emerge from temporally distant actions.
    \item We propose a history-aware graph repair framework, integrating Version Control-based error tracing, Edge Impact scoring.
    \item We refine the MANGO \citep{ding2024mangobenchmarkevaluatingmapping} benchmark dataset by systematically removing all non-topological actions and many pre-existing structural inconsistencies, creating a cleaned variant whose LLM-built input graphs provide a controlled testbed for map repair.
\end{itemize}

\subsection{Related Work}
\paragraph{Enhancing the Spatial Reasoning Ability of LLMs.}
In recent years, the spatial reasoning ability of large language models (LLMs) has been improved through specialized training and prompting strategies. AlphaMaze~\citep{dao2025alphamaze} combines supervised learning with reinforcement learning (GRPO) for maze navigation, while Mind's Eye~\citep{wu2024mind} introduces a ``visualization-of-thought'' prompting technique to simulate internal spatial representations. Although these approaches achieve progress in reasoning, they remain constrained by the context window and lack mechanisms to maintain long-term consistency during extended spatial reasoning tasks.

\paragraph{Mapping Evaluation in Language Agents.}
LLMs demonstrate a certain degree of spatial understanding when reasoning within short contexts, but they encounter significant limitations in long-horizon or complex textual reasoning. The MANGO benchmark~\citep{ding2024mangobenchmarkevaluatingmapping} shows that even GPT-4, due to its context length limitation, can only process the first 70 steps of a text-based environment. To mitigate this, modular navigation frameworks~\citep{zhang2025llmnav} introduce planning–execution modules. However, they lack graph-level consistency tracking mechanisms, which causes structural errors to accumulate unnoticed during extended exploration. These findings highlight that \textit{localized or incremental mapping} is a necessary pathway for extending LLM reasoning in complex environments. To further address these limitations, it is instructive to look at Simultaneous Localization and Mapping (SLAM), where incremental mapping and long-term coherence have been extensively studied.

\paragraph{SLAM as Inspiration.}
SLAM methods provide theoretical foundations for incremental mapping and consistency maintenance. GraphSLAM~\citep{lu1997globally} formulates mapping as graph optimization, while loop closure techniques~\citep{galvez2012bags,cummins2008fab} detect revisited places to trigger global corrections. ORB-SLAM~\citep{mur2015orb} and Cartographer~\citep{hess2016cartographer} demonstrate effective consistency maintenance by separating local and global optimization.

\paragraph{Error Management.}
While SLAM methods rely on robust kernels and outlier filtering, knowledge graph systems~\citep{zhang2024oneeditneuralsymboliccollaborativelyknowledge,lu2025karmaleveragingmultiagentllms,gil2024convergconcurrentversioningknowledge} have developed explicit conflict resolution and version control mechanisms. These approaches demonstrate that maintaining coherence during incremental updates requires systematic error detection beyond geometric constraints.

\paragraph{Incremental Scene Graph Construction.}
Scene graph methods~\citep{wu2021scenegraphfusionincremental3dscene,gu2024conceptgraphs,yin2024sgnav} explore incremental semantic mapping through multi-view fusion and LLM-based relation inference. However, most lack systematic mechanisms for conflict detection and consistency maintenance during construction.

Building on SLAM's structural principles, we address logical conflicts from LLM reasoning through \textit{version control} that records complete reasoning history and \textit{Edge Impact Scoring} that enables targeted error localization and rollback.

\section{Approach}
\subsection{Overview of the Framework}
\begin{figure*}[t]
  \centering
  \includegraphics[width=\linewidth]{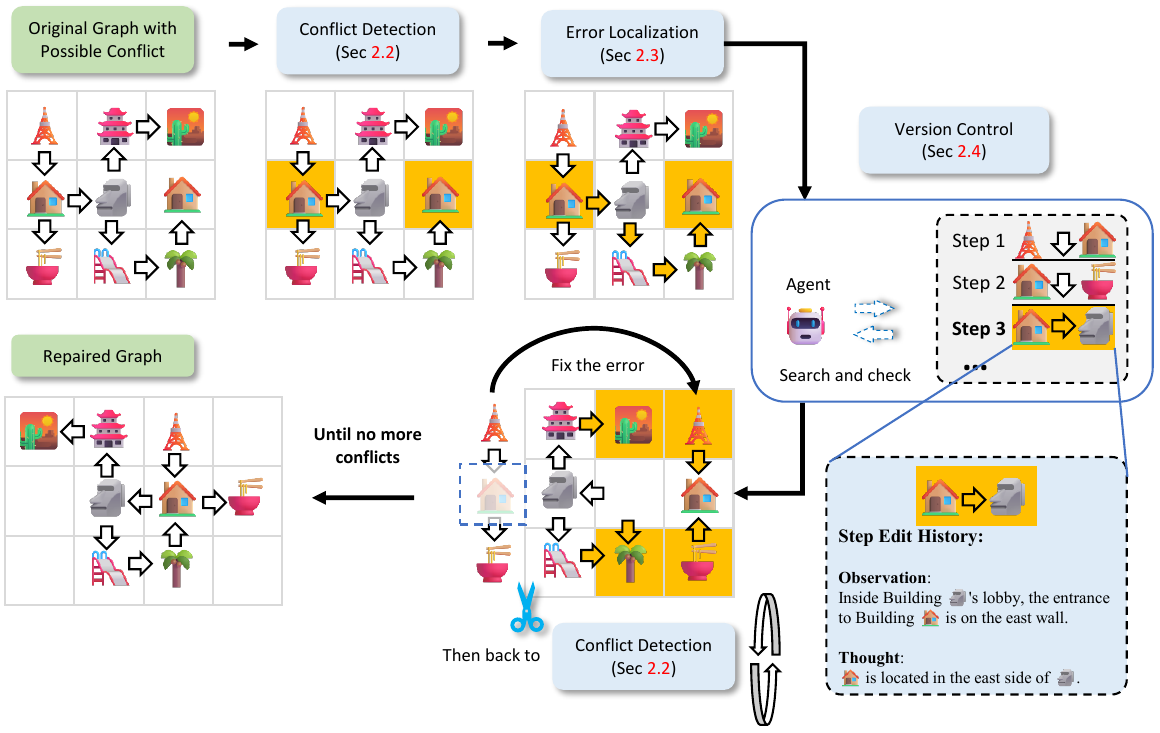}
  \caption{Overview of the LLM-MapRepair framework.}
  \label{fig:framework}
\end{figure*}

As mentioned in Sec.~\ref{introduction}, directly using LLM context to process complex textual scenarios is impractical due to context length limitations~\citep{ding2024mangobenchmarkevaluatingmapping}. We introduce an incremental graph construction approach where the LLM incrementally updates a graph by recording spatial relationships from each new observation. To ensure structural consistency in partially observable environments, we propose a modular repair framework that detects and corrects errors as they emerge during exploration.

Figure \ref{fig:framework} illustrates the complete workflow of the LLM-MapRepair framework for spatial graph construction and repair. The framework operates cyclically to transform conflict-laden graphs into structurally consistent representations through systematic error detection and targeted corrections.

The repair process consists of three stages: (1) Conflict Detection (Sec.~\ref{conflict_detection}) identifies structural inconsistencies; (2) Error Localization (Sec.~\ref{error_localization}) employs Edge Impact Scoring to prioritize erroneous edges and trace their origins; (3) Version Control (Sec.~\ref{version_control}) maintains historical context, preserving observations and reasoning processes. When examining commit history, the agent accesses complete contextual information to identify flaws and apply targeted repairs. The system iterates until the graph becomes conflict-free or the repair budget (outer iterations and per-conflict attempts) is exhausted, which is especially effective when early errors manifest much later in exploration.

\subsection{Conflict Detection}
\label{conflict_detection}

As LLM agents build navigation graphs from text, inconsistencies may gradually accumulate, resulting in structural conflicts. We identify three major types---\textit{naming}, \textit{directional}, and \textit{topological}---as illustrated in Figure~\ref{fig:conflict}(a). In our cleaned-MANGO evaluation, direction conflicts dominate in the initial LLM-mapped graphs; the framework detects all three types, which is necessary for downstream natural-language deployments such as the DRC case study where naming overlaps can occur whenever the front-end LLM uses ambiguous location strings:

\begin{figure*}[t]
\centering
\begin{subfigure}[b]{0.38\textwidth}
    \centering
    \includegraphics[width=\linewidth]{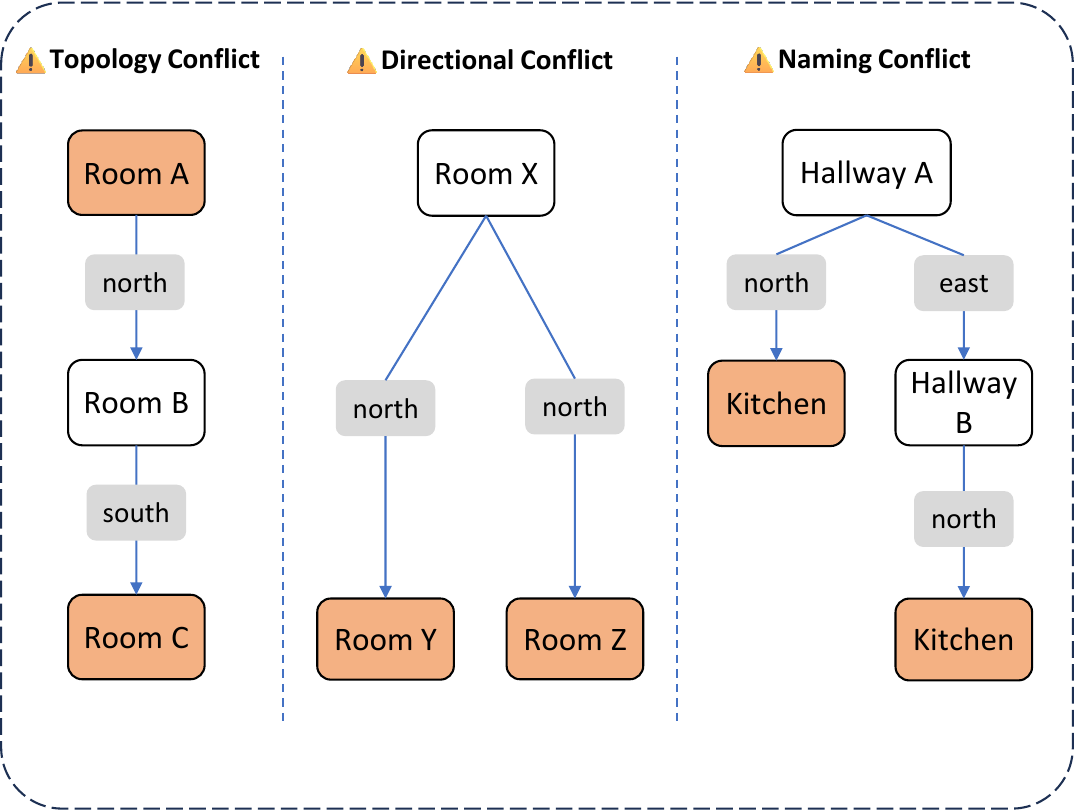}
    \caption{Conflict Types}
    \label{fig:conflict-types}
\end{subfigure}
\hfill
\begin{subfigure}[b]{0.6\textwidth}
    \centering
    \includegraphics[width=\linewidth]{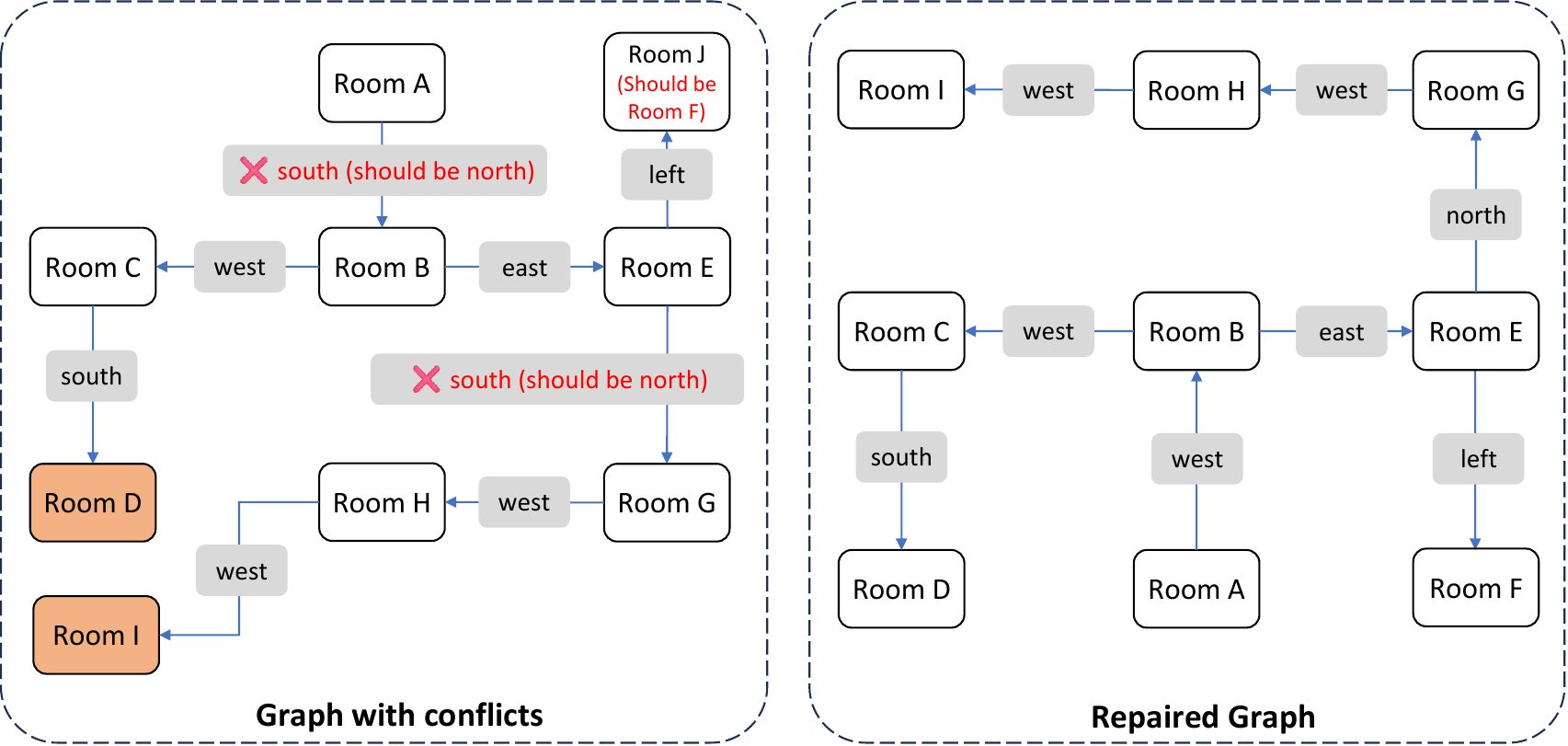}
    \caption{Challenge Scenario}
    \label{fig:challenge}
\end{subfigure}
\caption{
(a) Three types of structural conflicts — naming conflict, directional conflict, and topological conflict. Highlighted nodes indicate the conflicting pairs.
(b) A challenging conflict scenario: a misdirected edge from Room E to Room G introduces a latent spatial misalignment. A visible topology conflict emerges later between Room D and Room I (highlighted), while correcting the edge E$\rightarrow$G may trigger a new conflict between Room H and Room A. The edge E$\rightarrow$J is incorrect, though it is unrelated to the current conflict.
}
\label{fig:conflict}
\end{figure*}

\begin{itemize}
\item \textbf{Topological Conflict:} Invalid graph structures such as cycles in tree-like spaces, unreachable nodes, or over-connected components.

\item \textbf{Directional Conflict:} Multiple outgoing edges with the same direction label violating spatial constraints.

\item \textbf{Naming Conflict:} Identical names assigned to different locations, causing ambiguity in reasoning and localization.
\end{itemize}

\subsection{Error Localization}
\label{error_localization}
Resolving conflicts in LLM-generated navigation graphs is often more difficult than detecting them, due to several intertwined challenges illustrated in Figure~\ref{fig:conflict}(b). The primary complexity stems from \textbf{delayed conflicts}, where errors introduced early may not be noticed until much later in the exploration process. As demonstrated in the challenge scenario, a wrong direction from \textit{Room~E to G} leads to a cascade of misplacements, but the actual conflict only becomes apparent when the loop reaches overlapping \textit{Rooms~D and I}. This temporal gap between error introduction and detection is further complicated by \textbf{entangled conflicts}, where attempting to fix one edge can inadvertently create new conflicts elsewhere in the graph. For instance, adjusting the \textit{Room~E$\rightarrow$G} connection resolves the initial overlap but simultaneously causes a new conflict between \textit{Rooms~H and A}. Perhaps most problematically, \textbf{silent errors} can persist undetected due to the absence of contradictory evidence. The incorrect direction from \textit{Room~E to J} exemplifies this issue, causing no immediate conflict while silently corrupting the underlying map structure.

The asynchrony between graph construction errors and structural conflicts motivate our framework's separation of \textbf{conflict detection} and \textbf{error localization}, enabling robust identification of true error sources through temporal and structural reasoning.

Once a conflict is detected, the system must identify not just the conflicting edges, but the actual root cause that introduced the inconsistency. This is non-trivial, as the erroneous edge may lie far from the observed conflict and may even appear structurally correct in isolation. Our localization pipeline proceeds in four stages: (1) identifying the minimal conflicting path pair, (2) computing their lowest common ancestor (LCA), (3) extracting divergent edges as error candidates, and (4) scoring and ranking these candidates by impact.

\paragraph{Minimal Conflicting Path Pair}
Given a structural conflict (e.g., naming or topology), we first locate two distinct paths that lead to the conflicting nodes. For instance, in Figure~\ref{fig:conflict} challenge scenario, the topology conflict between Room D and Room I can be traced to two paths:
\begin{align*}
\text{Path}_1 &: \text{Room B} \rightarrow \text{C} \rightarrow \text{D} \\
\text{Path}_2 &: \text{Room B} \rightarrow \text{E} \rightarrow \text{G} \rightarrow \text{H} \rightarrow \text{I}
\end{align*}
Both paths result in overlapping node positions, violating spatial exclusivity constraints.

\paragraph{Lowest Common Ancestor (LCA)}
To identify where the error first diverged, we compute the lowest common ancestor (LCA) of the two conflicting paths. This is the \emph{last} node shared between them before the divergence that leads to inconsistency.

\textbf{Domain Clarification.} Critically, LCA is computed not over the spatial graph $G$ (which may contain cycles and loops), but over the \emph{Reasoning History Tree} $\mathcal{T}$—a directed acyclic graph (DAG) encoding the temporal dependency structure of edge additions during graph construction. Each node $v \in \mathcal{T}$ corresponds to a commit in the version control system and is associated with a unique timestamp $\tau(v)$ reflecting when the corresponding edge was introduced. This temporal ordering ensures that the LCA is well-defined even when the spatial graph contains cycles.

Formally, for two conflicting paths $p_1$ and $p_2$ in the spatial graph $G$, we trace their corresponding reasoning paths $\pi_1, \pi_2$ in $\mathcal{T}$ and compute:
\begin{equation}
\text{LCA}(\pi_1, \pi_2) = \operatorname*{arg\,max}_{v \in \pi_1 \cap \pi_2} \tau(v)
\end{equation}
where $\tau(v)$ denotes the timestamp or topological depth of node $v$ in $\mathcal{T}$. The use of $\operatorname*{arg\,max}$ ensures we identify the \emph{most recent} (temporally latest) common ancestor—the true divergence point where paths began to differ.

\textbf{Unified Logic for All Conflict Types.} This formulation provides a unified framework for both topological and directional conflicts:
\begin{itemize}
    \item \textbf{Long-range conflicts} (topology violations): The LCA is strictly earlier than the source node of the current observation ($\text{LCA} \neq \text{Source}$), indicating a \emph{latent error} introduced in a previous step. In Figure~\ref{fig:conflict}, Room~B serves as the LCA for the conflict between D and I, with the error lying along edges downstream from B.
    \item \textbf{Local conflicts} (directional violations): The LCA coincides with the source node itself ($\text{LCA} = \text{Source}$), representing a degenerate case or \emph{zero-length divergence}. This indicates an \emph{immediate error} where the current edge directly contradicts local geometric constraints (e.g., compass directions).
\end{itemize}
This distinction allows us to apply a common candidate-generation procedure to both conflict types. We use room names as shorthand in the running example; formally, the LCA is a commit node in $\mathcal{T}$, and each commit is mapped back to the spatial edge or room introduced at that step before candidate edges are extracted in $G$. Edges beyond the LCA node are then considered candidate error sources for subsequent ranking.

\paragraph{Candidate Edge Extraction}
We extract the divergent subpaths from the LCA to each conflict node and collect all edges along these subpaths as potential causes of the inconsistency. In the example:
\[
\begin{aligned}
\text{Candidate edges} = \{\text{B}\rightarrow\text{C},\ \text{C}\rightarrow\text{D},\ \text{B}\rightarrow\text{E}, \\ \text{E}\rightarrow\text{G},\ \text{G}\rightarrow\text{H},\ \text{H}\rightarrow\text{I}\}
\end{aligned}
\]
Additionally, silent errors (e.g., E$\rightarrow$J) not yet resulting in conflicts can also be included as fallback candidates for global ranking. In the Appendix B TC1 reconstruction of this scenario, the same procedure yields 8 of the graph's 9 edges (the divergent-subpath set above plus two conflict-adjacent fallback edges; the single edge filtered out is on an unrelated branch).

\paragraph{Edge Scoring and Ranking}

To determine which candidate edge to prioritize for inspection and repair, we assign each edge a composite score based on three factors: \textit{reachability}, \textit{conflict count}, and \textit{usage}. These reflect the potential structural impact, the degree of inconsistency evidence, and the reliance of observed paths on the edge, respectively.

Inspired by PageRank~\citep{brin1998anatomy}, we model edge importance through three factors:

\begin{itemize}
    \item \textbf{Reachability}: downstream nodes reachable from edge $e$, reflecting propagation potential.
    \item \textbf{Conflict Count}: distinct conflicts involving $e$, indicating contribution to inconsistencies.
    \item \textbf{Usage}: conflict-related paths including $e$, capturing empirical relevance.
\end{itemize}

We adopt an unweighted scoring function after min-max normalization:
\begin{equation}
\label{eq:scoring}
\text{score}(e) = \widehat{\text{Reach}}(e) + \widehat{\text{Conflict}}(e) + \widehat{\text{Usage}}(e)
\end{equation}
\paragraph{Repair Prioritization}
We prioritize edges by descending score to maximize conflict revelation—repairing high-impact edges first to either resolve existing conflicts or expose hidden errors. To trace an edge's origin or reverse mistaken fixes, we maintain temporal structure through Version Control.

\subsection{Version Control}
\label{version_control}
Version Control maintains a totally ordered sequence of commits, enabling targeted rollback, difference analysis, and complete reasoning history retrieval. The Reasoning History Tree $\mathcal{T}$ used by the LCA-based localizer (\S\ref{error_localization}) is the rooted DAG induced by these commits via their edge-source dependencies---commits form the chain, $\mathcal{T}$ records which commit introduced each spatial edge---so the two views describe the same underlying history at different granularities.

\paragraph{Version Control Structure.}
Version Control is a totally ordered sequence of version records $[G_0, G_1, \dots, G_t]$, where each commit $G_i$ represents a step-wise change to the graph. Rather than storing full graph snapshots, each version logs only the incremental updates (e.g., edge additions or conflict-triggered replacements), with metadata including the step identifier, specific edge changes (+ for additions, - for removals during replacements), trigger event type, and associated observation.

\[
\begin{aligned}
G_i = \{&\texttt{Step\_id},\ \texttt{Commit},\ \texttt{Trigger\_event},\\
&\texttt{Observation\_id},\ \texttt{Analysis}\}
\end{aligned}
\]
This structure minimizes memory cost while enabling exact reconstructions.

\paragraph{Supported Operations.}
Version Control supports three key operations:
\begin{itemize}
    \item \texttt{rollback\_to(version)}: Restores the graph to a prior state by undoing subsequent steps.
    \item \texttt{recall\_step(version)}: Obtain the thinking history corresponding to the step.
    \item \texttt{diff(G$_i$, G$_j$)}: Computes edge-level differences between two versions.
\end{itemize}
These operations support both runtime repair decisions and post hoc analysis.

\paragraph{Incremental Evolution.}
Every LLM-initiated interaction—whether through new observations or repair actions—triggers a graph update and logs a new version in Version Control. This guarantees that even failed or partially correct decisions are preserved for future analysis. Version Control records whether an update was conflict-triggered (e.g., \texttt{Trigger\_event = conflict\_repair}) to provide interpretability in version history.
Version Control enables following key capabilities:
\begin{itemize}
    \item \textbf{Graph alignment}: Compare versions before and after a repair to assess changes.
    \item \textbf{Structural diffing}: Detect which steps introduced regressions or inconsistencies.
    \item \textbf{Error propagation tracing}: Model how errors spread over time.
\end{itemize}
Unlike flat logs or event lists, the versioned graph design supports non-destructive rollbacks and dependency-aware repair strategies.

This aligns with well-established principles in database systems, where \textit{write-ahead logging (WAL)}~\citep{wal} ensures recoverability and traceability by recording all state changes. Similarly, Version Control gives LLM-based systems a foundation for provenance tracking, self-debugging, and structured repair---all essential for interactive, persistent reasoning tasks.

\section{Experiment}
\subsection{Datasets and Evaluation Setup}
\label{sec:datasets}

We evaluate the framework on four evaluation settings spanning controlled-error, procedurally-generated, and natural-language settings.

\paragraph{Synthetic random graphs (ablation).} For the per-component ablation in Table~\ref{tab:repair_summary}, we generate random directed graphs of $|V|=60$ nodes with seed-controlled edge sampling. Conflicts are injected by selecting $k\in\{4, 8\}$ ground-truth edges uniformly at random and rewriting each either by perturbing its direction label (\emph{direction} error) or by replacing its target with a non-adjacent node that produces a topological overlap (\emph{topology} error). Ground truth is preserved separately, enabling per-mode comparison; each $\langle$conflict type, density$\rangle$ cell aggregates 20 independent seeds.

\paragraph{Synthetic with direction-conflict noise (cross-vendor).} For cross-vendor generalization on a fully observable substrate (left half of Table~\ref{tab:method_vs_baseline}), we generate synthetic graphs in the same family as the ablation set and inject direction-conflict noise only (1 to 3 conflicts per graph). Each cell aggregates $n{=}20$ seeds.

\paragraph{TextWorld procedurally-generated games.} For cross-vendor generalization on a real-language substrate (right half of Table~\ref{tab:method_vs_baseline}), we use TextWorld~\citep{cote2018textworld}, which procedurally generates text-adventure games with diverse room layouts and natural-language room descriptions. We inject a noise mixture comprising three failure modes derived empirically from LLM outputs on interactive-fiction walkthroughs: room-name collapses, duplicate-direction edges, and hallucinated rooms. Each cell aggregates 30 game-seeds.

\paragraph{Real IF maps from MANGO walkthroughs.} For cross-vendor evaluation on graphs that originate from natural-language input rather than direct noise injection (Table~\ref{tab:mango_frontier}), we use the maps produced when gpt-4.1 incrementally constructs a navigation graph from the cleaned MANGO walkthroughs. We evaluate on all 42 cleaned-MANGO games whose resulting input graphs contain at least one residual conflict (534 conflicts in aggregate), avoiding any sampling bias from selecting a sub-population. We then run baseline, Edge-Impact, and VC+EI repair with three models from three vendors (GPT-5.5, Claude-Haiku~4.5, Gemini~3.5-Flash) via the openai-hub proxy; we evaluate one representative model per vendor here to keep repair-stage cost tractable on the 42-game suite, with the remaining four Table~\ref{tab:method_vs_baseline} models still covered on the cheaper synthetic and TextWorld substrates. We additionally report two non-LLM references (\texttt{heuristic\_remove} and \texttt{heuristic\_modify}) for context.

\paragraph{Real-text deployment (\emph{Dream of the Red Chamber}).} For end-to-end deployment on natural text, we evaluate on Chapters~16 and~17 of the Chinese classical novel \emph{Dream of the Red Chamber}, with a manually annotated ground-truth map (35 unique locations and 34 spatial relation pairs, evaluated as undirected pairs rather than directed arcs, authored by a domain expert). The LLM front-end incrementally constructs a navigation graph from chapter prose by processing paragraph-level descriptions sequentially; each detected location becomes a node and each described transition becomes an edge. The repair framework then resolves accumulated structural inconsistencies. Quantitative recall metrics are reported in Table~\ref{tab:drc_recall} and the resulting map is visualized in Figure~\ref{fig:problem-overview}.

A separate dataset contribution is a cleaned variant of the MANGO benchmark~\citep{ding2024mangobenchmarkevaluatingmapping}, which removes non-topological actions and many of the pre-existing structural inconsistencies from the original 53 interactive-fiction environments; the 6-step refinement pipeline is detailed in Appendix~\ref{sec:appendix_dataset}.

\paragraph{Hyperparameters and metrics.} Across all LLM-driven experiments, the repair agent is permitted up to 20 outer iterations and 3 attempts per individual conflict before the conflict is marked as unresolved. The reported \emph{Conflict-Free (CF)} repair rate is the fraction of graphs whose final state, after repair termination, contains zero residual conflicts under the detector defined in \S\ref{conflict_detection}. The choice of $20$ outer iterations was validated by a sensitivity sweep on 5 representative cleaned-MANGO games with GPT-5.5\,+\,EI: aggregate net resolution rate (repaired minus newly-introduced, normalized by input conflicts; positive values indicate net repair) are $-30.6\%$, $-32.4\%$, $\mathbf{+50.9\%}$, and $-17.6\%$ at \texttt{max\_iter} $\in\{5,10,20,40\}$ respectively. Below 20 the agent runs out of budget on graphs that need multi-edge rewriting; above 20 the additional budget is spent on stochastic exploration that introduces new conflicts faster than they are resolved, so 20 strikes the best aggregate trade-off.

\begin{table}[t]
\centering
\caption{Per-component ablation on synthetic graphs. Conflict-free (CF) repair rate (\%) with $95\%$ Wilson confidence interval by conflict type and error density on random graphs of size 60, with gpt-4.1 as the repair LLM and $n{=}20$ independent seeds per cell. \textbf{Base.}, unscaffolded LLM; \textbf{EI}, Edge-Impact Ranking; \textbf{VC}, Version Control; \textbf{VC+EI}, their combination.}
\small
\setlength{\tabcolsep}{3pt}
\begin{tabular}{llcccc}
\toprule
\textbf{Conflict} & \textbf{Err.} & \textbf{Base.} & \textbf{EI} & \textbf{VC} & \textbf{VC+EI} \\
\midrule
Topology  & 4 & 50 \scriptsize{(30,70)} & \textbf{95 \scriptsize{(76,99)}} & 50 \scriptsize{(30,70)} & 50 \scriptsize{(30,70)} \\
Topology  & 8 & 30 \scriptsize{(15,52)} & \textbf{60 \scriptsize{(39,78)}} & 25 \scriptsize{(11,47)} & 40 \scriptsize{(22,61)} \\
Direction & 4 & 70 \scriptsize{(48,86)} & 75 \scriptsize{(53,89)} & 55 \scriptsize{(34,74)} & 55 \scriptsize{(34,74)} \\
Direction & 8 & 70 \scriptsize{(48,86)} & 50 \scriptsize{(30,70)} & 60 \scriptsize{(39,78)} & 25 \scriptsize{(11,47)} \\
\bottomrule
\end{tabular}
\label{tab:repair_summary}
\end{table}

\begin{table}[t]
\centering
\caption{Generalization across LLMs from OpenAI, Anthropic, and Google. Conflict-Free (CF) repair rate (\%) on (i) synthetic graphs with direction-conflict noise (n=20 per cell) and (ii) TextWorld procedurally-generated text-adventure games with a mango-like noise mixture (n=30 per cell). Bold entries denote configurations where VC+EI outperforms the baseline LLM in that cell.}
\small
\setlength{\tabcolsep}{3pt}
\begin{tabular}{lcccc}
\toprule
\multirow{2}{*}{\textbf{Model}} & \multicolumn{2}{c}{\textbf{Synthetic CF (\%)}} & \multicolumn{2}{c}{\textbf{TextWorld CF (\%)}} \\
\cmidrule(lr){2-3} \cmidrule(lr){4-5}
 & \textbf{Base} & \textbf{Ours} & \textbf{Base} & \textbf{Ours} \\
\midrule
GPT-5.5             & 25.0 & \textbf{75.0} & 20.0 & 20.0          \\
GPT-5-mini          & 20.0 & \textbf{35.0} & 16.7 & 16.7          \\
o4-mini             & 30.0 & 20.0          & 20.0 & \textbf{26.7} \\
Claude-Sonnet~4.6   & 30.0 & \textbf{40.0} & 20.0 & \textbf{33.3} \\
Claude-Haiku~4.5    & 10.0 & \textbf{30.0} & 16.7 & \textbf{33.3} \\
Gemini~2.5-Flash    & 25.0 & 10.0          & 20.0 & \textbf{26.7} \\
Gemini~3.5-Flash    & 20.0 & \textbf{50.0} & 20.0 & \textbf{33.3} \\
\bottomrule
\end{tabular}
\label{tab:method_vs_baseline}
\end{table}

\begin{table}[t]
\centering
\caption{Per-component repair on real IF maps. Residual conflicts (lower is better) after repair on all 42 cleaned-MANGO games whose gpt-4.1-built input graphs contained at least one residual conflict (534 input conflicts in aggregate). Counts above 534 indicate that the corresponding repair mode introduced additional conflicts during repair. The two non-LLM references at the bottom bracket the LLM modes: \texttt{heuristic\_remove} aggressively deletes any conflict-participating edge (strongest absolute reducer); \texttt{heuristic\_modify} relabels directions instead (weaker; in our evaluation it also has the largest ground-truth edge loss, see Appendix Table~\ref{tab:mango_preservation}). Best LLM cell per model is bolded.}
\footnotesize
\setlength{\tabcolsep}{4pt}
\begin{tabular}{lccc}
\toprule
\textbf{Method} & \textbf{GPT-5.5} & \textbf{Claude-H~4.5} & \textbf{Gemini-3.5F} \\
\midrule
Baseline LLM      & 609 & 874 & 841 \\
EI                & \textbf{396} & 827 & \textbf{572} \\
VC+EI             & 458 & \textbf{625} & 837 \\
\midrule
\multicolumn{4}{l}{\emph{Non-LLM references (model-independent)}} \\
heuristic\_modify & \multicolumn{3}{c}{438} \\
heuristic\_remove & \multicolumn{3}{c}{98} \\
\bottomrule
\end{tabular}
\label{tab:mango_frontier}
\end{table}

\begin{table}[t]
\centering
\caption{End-to-end LLM-MapRepair on natural text. Ground-truth-matched node and edge recall (\%) on Chapters~16--17 of \emph{Dream of the Red Chamber} against a 35-node, 34-relation human-authored map. \textbf{Baseline LLM} is direct incremental construction without our repair loop; \textbf{LLM-MapRepair} adds Conflict Detection + Edge-Impact Ranking + Version Control. Both use gpt-4.1. ``\#N''/``\#E'' are the predicted node/edge counts; higher recall is paired with higher false-positive node/edge counts, reflecting the discretization-driven over-generation discussed in \S\ref{limitations_sec}.}
\small
\setlength{\tabcolsep}{4pt}
\begin{tabular}{lcccc}
\toprule
\textbf{Method} & \textbf{\#N} & \textbf{\#E} & \textbf{Node R (\%)} & \textbf{Edge R (\%)} \\
\midrule
Baseline LLM        & 47  & 49  & 85.7 & 32.4 \\
\textbf{LLM-MapRepair} & 143 & 144 & \textbf{94.3} & \textbf{88.2} \\
\midrule
$\Delta$            & +96 & +95 & $+8.6$~pp & $+55.8$~pp \\
\bottomrule
\end{tabular}
\label{tab:drc_recall}
\end{table}

\subsection{Ablation Study}

To evaluate the contribution of each component in our repair framework, we run a controlled ablation on random synthetic graphs (size~60) with directly injected topology or direction conflicts at densities of 4 and 8 errors per graph; gpt-4.1 serves as the repair LLM and each cell of Table~\ref{tab:repair_summary} aggregates 20 independent seeds. We compare four settings: a baseline LLM without scaffolding (\textbf{Base.}); Edge-Impact Ranking Only (\textbf{EI}), which prioritizes repair candidates by their structural impact score; Version Control Only (\textbf{VC}), which exposes commit history and rollback as agent tools; and the combined Version Control + Edge-Impact Ranking (\textbf{VC+EI}).

Edge-Impact Ranking dominates topology conflicts, where the LCA structure is most informative. At 4 errors, EI reaches 95.0\% CF, compared to 50.0\% for the baseline and for VC alone; the advantage persists at higher density (60.0\% vs. 30.0\% at 8 errors). The Edge-Impact priority signal lets the agent identify and modify the high-cascade edge first, preventing the local-fix cascades observed in the unprioritized modes. On direction conflicts at moderate density, EI yields a smaller lift (75.0\% vs. 70.0\% baseline at 4 errors), reflecting the more direct nature of directional disambiguation, where local LLM reasoning is already competitive. At very high direction-conflict density (8 simultaneous errors), VC+EI drops to 25\% CF: under this regime the rollback machinery from Version Control accumulates inconsistent commit histories faster than the priority signal can resolve them, and the agent spends iterations un-doing partial rewrites it just committed.

Version Control complements Edge-Impact Ranking on two distinct axes: it preserves edge-level provenance for auditable repair and supports rollback during multi-step rewriting. Empirically, the value of the combined VC+EI configuration is model-dependent (Table~\ref{tab:mango_frontier}). For practitioners deploying our framework with an unseen repair LLM, we therefore recommend running both EI and VC+EI on a small held-out subset (e.g., $5$--$10$ conflict-bearing graphs) and selecting the mode that yields the lower residual count on that subset; this diagnostic costs at most a few minutes of LLM inference and reliably reproduces the per-model preference observed across our evaluation. EI alone is the appropriate default when no validation budget is available.

End-to-end deployment on the \emph{Dream of the Red Chamber} chapters illustrates that these scaffolds carry to natural text in a recall-oriented sense: applying LLM-MapRepair to the incrementally constructed map raises node recall from $85.7\%$ to $94.3\%$ ($+8.6$~pp) and edge recall from $32.4\%$ to $88.2\%$ ($+55.8$~pp) against the human-authored ground-truth map (Table~\ref{tab:drc_recall}), with the largest gain on edges where temporally distant traversals must be reconciled. The recall gains come with substantial over-generation---the predicted node and edge counts reach roughly $4\times$ the ground-truth counts (i.e.\ about $3\times$ more nodes/edges than the ground truth)---so the DRC result should be read as a recall-side improvement on a single-novel deployment rather than a balanced precision/recall claim.

The complementary pattern carries over to natural-language inputs. Evaluated on all 42 cleaned-MANGO games whose gpt-4.1-built input graphs contain non-zero conflicts (534 conflicts in aggregate), the best LLM configuration depends on the model: GPT-5.5 and Gemini~3.5-Flash favor EI alone ($609\!\to\!396$ and $841\!\to\!572$), whereas Claude-Haiku~4.5 benefits most from VC+EI ($874\!\to\!625$), where the commit-history signal compensates for that model's tendency to introduce new conflicts during local rewriting (Table~\ref{tab:mango_frontier}). Among LLM modes, the best configuration always strictly improves over the unscaffolded baseline; the cross-model improvement of the best LLM configuration over baseline is $35\%$ for GPT-5.5, $28\%$ for Claude-Haiku, and $32\%$ for Gemini. The two non-LLM references bracket the LLM range and expose a clear repair-versus-preservation trade-off. \texttt{heuristic\_remove} achieves the strongest absolute residual reduction ($98$ conflicts) by deleting any conflict-participating edge, at the cost of losing on average $-56$ ground-truth-correct edges; \texttt{heuristic\_modify} ($438$) trades that aggressive deletion for direction relabels, losing more ground-truth structure ($-97$ edges) when relabels are wrong. The LLM modes occupy intermediate points along this trade-off: Claude-Haiku~4.5 modes lose only $-31$ to $-67$ ground-truth-correct edges (best structural preservation among LLM-driven repair), whereas GPT-5.5 modes resolve more conflicts at the cost of higher edge loss ($-86$ to $-89$); the full per-method preservation table is given as Appendix Table~\ref{tab:mango_preservation}. On the headline repair metric, no LLM mode matches \texttt{heuristic\_remove}'s absolute residual reduction; LLM-driven repair is preferable when downstream tasks require the more structure-preserving repairs that LLM judgement provides, especially when the front-end has uncertain confidence over which edge in a conflict is the true error.

To verify the generalization of our framework across LLM vendors, we additionally embed \textbf{Version Control + Edge-Impact Ranking} into seven LLMs spanning three vendors (OpenAI, Anthropic, Google); see Table~\ref{tab:method_vs_baseline}. On synthetic direction-conflict repair, our method provides positive lift on five of the seven models, with GPT-5.5 showing the largest single-model improvement of $+50.0$~pp ($25.0\%\!\to\!75.0\%$) and Gemini~3.5-Flash a further $+30.0$~pp ($20.0\%\!\to\!50.0\%$); two models (o4-mini and Gemini~2.5-Flash) show small regressions, which we attribute to two factors: (i) both models already achieve a relatively high baseline (30.0\% and 25.0\% respectively), reducing the headroom available for an additional priority signal to help; and (ii) under our system prompt, both models exhibit a tendency to delete edges defensively when given Edge-Impact scores, occasionally removing a correct edge whose downstream subgraph is later flagged as inconsistent. On TextWorld procedurally-generated text-adventure games with a mango-like noise mixture (n=30 per cell), five of the seven models show positive lift over the baseline LLM, led by Claude-Haiku~4.5 ($+16.7$~pp), Claude-Sonnet~4.6 ($+13.3$~pp), and Gemini~3.5-Flash ($+13.3$~pp). The absolute TextWorld CF rates are bounded by the procedurally-generated layouts (no-repair floor at $16.7\%$), so relative lift over baseline is the appropriate comparison. These results suggest that the structural scaffolding contributed by VC+EI provides measurable gains on the majority of model--task cells we evaluated, particularly on real-language substrates; statistical significance is not formally assessed at the per-cell sample sizes ($n{=}20$ synthetic / $n{=}30$ TextWorld).

\subsection{Algorithmic Validation without LLM}
To isolate and validate the core algorithmic contributions, namely LCA based candidate filtering and edge impact scoring, independently of LLM performance variability, we design a controlled experimental suite on synthetic graphs with known ground truth errors, as detailed in Appendix~\ref{sec:appendix_algo_validation}. Unlike the LLM-driven experiments above, this setup directly injects topological, directional, and naming conflicts into constructed graphs, enabling precise evaluation without confounding effects from language understanding or generation quality.

\paragraph{Key Findings.}
Across six synthetic scenarios, LCA based filtering reduces the candidate edge search space by an average of 24.6\%, with reductions of up to 75\% in directional conflicts where the LCA coincides with the source node, corresponding to the degenerate case discussed in \S\ref{error_localization}. Edge impact scoring is consistent with true cascade potential at the smallest non-trivial scale (Spearman $\rho = 1.0$ on the 5-edge sanity-check scenario; treated as illustrative rather than statistically robust). Priority based inspection further accelerates high impact error discovery by a factor of 2.3 compared to random traversal, requiring 56.5\% fewer edge examinations to identify critical errors. These results illustrate that the proposed mechanisms behave as intended at the algorithmic level, independent of LLM capabilities; a larger-suite scaling experiment (Appendix~\ref{sec:appendix_algo_validation}) supplies a broader counterpart for the LCA-filtering component on 1{,}160 programmatically generated graphs.

\section{Conclusion}
Although LLMs can incrementally construct topological maps from language observations, their outputs are often noisy, accumulating misaligned edges, duplicates, and latent inconsistencies that degrade downstream reasoning. We address this with a three-stage repair pipeline comprising conflict detection, error localization, and impact-aware correction. A version-controlled map history enables rollback, difference analysis, and causal tracing of errors, while an Edge Impact Score prioritizes high-cascade edges for inspection by quantifying each edge's structural and usage-based influence.

\section{Limitations}
\label{limitations_sec}
Our approach has several limitations. First, the synthetic and TextWorld substrates used for ablation and cross-vendor evaluation rely on discrete cardinal directions and unit distances, which enables precise symbolic conflict detection. However, real-world novels contain continuous spatial descriptions with arbitrary angles and imprecise distances, where conflicts may be obscured by accumulated errors. While our discretization approach (binning angles into directional categories and distances into coarse ranges with elastic optimization) shows promise on literary texts such as \textit{Dream of the Red Chamber} (Figure~\ref{fig:problem-overview}), it substantially increases false positive rates due to approximate granularity. Second, our current conflict detection relies on observable structural violations and cannot identify latent semantic inconsistencies that do not manifest as topological errors. Third, the edge impact scoring heuristic, while effective empirically, lacks theoretical guarantees for optimal repair ordering in all graph configurations. Fourth, the end-to-end natural-text deployment in Table~\ref{tab:drc_recall} is demonstrated on a single novel (\emph{Dream of the Red Chamber}, Chapters~16--17) with one repair LLM (gpt-4.1); the algorithmic mechanism and cross-vendor robustness are corroborated separately on Tables~\ref{tab:method_vs_baseline} and~\ref{tab:mango_frontier}, but a broader literary corpus and additional repair-LLM vendors on the same end-to-end pipeline remain future work. Fifth, the structural priority signal from Edge-Impact Ranking is most informative on topology-driven conflicts. On direction conflicts at very high error density (eight or more simultaneous errors on a single graph), each conflicting edge admits multiple plausible direction labels with similar local impact, and the priority signal degrades: an unscaffolded LLM that processes conflicts sequentially can match or exceed the scaffolded modes in this regime. These limitations suggest important directions for future work in continuous spatial reasoning, semantic conflict modeling, and provably optimal repair strategies.

\section{Ethical Considerations}
In preparing this manuscript, LLMs were used only for language polishing and copy-editing; they did not contribute to the conceptualization, experimental design, analysis, or conclusions of the work.

\bibliography{references}

\appendix
\renewcommand{\thefigure}{A\arabic{figure}}
\renewcommand{\thetable}{A\arabic{table}}
\setcounter{figure}{0}
\setcounter{table}{0}

\section{Dataset Refinement Details}
\label{sec:appendix_dataset}
\subsection{Dataset Refinement}
 We analyze the MANGO trajectories and find that the dataset itself contains structural inconsistencies, which we categorize into \textit{directional conflicts}, and \textit{topological conflicts}.

\paragraph{Original Dataset Conflicts.} The original MANGO dataset \citep{ding2024mangobenchmarkevaluatingmapping} releases 53 interactive-fiction environments grouped under 18 parent game titles; every environment contains various types of conflicts as released, including many non-topological actions such as ``pray'' that do not correspond to spatial movement.

\paragraph{Refined Dataset Creation.} To better evaluate the performance of our LLM-MapRepair system, we create a refined dataset that removes pre-existing inconsistencies introduced by non-topological actions and retains only topologically meaningful navigation actions. Our dataset refinement process follows a systematic 6-step pipeline:

\begin{enumerate}
\item \textbf{Action Filtering}: Filter the original dataset to retain only 14 topological movement actions (north, south, east, west, up, down, northeast, northwest, southeast, southwest, in, out, enter, exit), removing non-spatial actions like ``pray''.

\item \textbf{Directional Conflict Resolution}: Remove duplicate edges when the same source node has multiple outgoing edges with identical direction labels. We preserve the edge with the minimum step count and remove all other duplicates.

\item \textbf{Topological Conflict Resolution}: Eliminate inconsistent reverse edges when bidirectional connectivity violates spatial symmetry. Edges that create directional mismatches in their reverse direction are removed.

\item \textbf{Reverse Edge Conflict Resolution}: Remove original edges that would cause directional conflicts when their corresponding reverse edges are added to maintain graph symmetry.

\item \textbf{Naming Conflict Resolution}: Eliminate edges that cause indirect conflicts through transitive spatial relationships. When different paths lead to inconsistent positional inferences for the same location, edges in the indirect paths are removed.

\item \textbf{Self-Loop Removal}: Delete all self-referential edges where a node points to itself, as these do not represent meaningful spatial transitions.
\end{enumerate}

This refinement process removes a total of 160 edges, reducing the edge count from 1,673 to 1,513. The resulting refined dataset (\texttt{data\_fixed}) retains edges that represent valid topological relationships with clear spatial semantics, and serves as the ground-truth reference for the LLM mapping and repair experiments. The 42 conflict-bearing input graphs used in Table~\ref{tab:mango_frontier} aggregate to 534 conflicts introduced when gpt-4.1 maps the refined walkthroughs into navigation graphs.

\paragraph{From Grid-Based Games to Real-World Novels.}
Our refined MANGO dataset uses standardized cardinal directions (north, south, east, west, etc.) and unit distances, which simplifies conflict detection through discrete symbolic matching. However, real-world novels present continuous spatial descriptions with arbitrary angles and imprecise distances. In such settings, conflicts may be obscured by accumulated continuous errors, and novels often lack precise angular or distance information altogether.

To bridge this gap, we employ a discretization approach: angles are binned into finite directional categories, and distances are partitioned into coarse ranges (e.g., near, medium, far). We further apply elastic optimization based on conflict severity, allowing tolerance thresholds to adapt dynamically. Under appropriate configurations, this approach improves mapping success for real novels. For instance, Figure~\ref{fig:problem-overview} demonstrates our system's performance on chapters 16--17 of \textit{Dream of the Red Chamber}, achieving node recall of $94.3\%$ (an $8.6$ percentage point improvement) and edge recall of $88.2\%$ (a $55.8$ percentage point improvement). However, the discretization strategy also increases false positive rates, as approximate binning may introduce spurious conflicts or miss genuine errors masked by coarse granularity.

\paragraph{Topological Conflicts.} These occur when an edge violates expected spatial symmetry or consistency. For example, in dataset \texttt{zork2}, a topological conflict is triggered when the system observes \texttt{ledge in ravine -- down --> deep ford}, whereas the expected reverse transition from earlier was \texttt{deep ford -- north --> ledge in ravine}. Based on this prior observation, the forward edge should have been labeled \texttt{south}, not \texttt{down}.

\paragraph{Directional Conflicts.} These arise when a node has multiple outgoing edges with the same direction label. For example, in dataset \texttt{zork2}, the node \texttt{carousel room} has two edges labeled ``north'', pointing to both \texttt{marble hall} and \texttt{topiary}. This violates the uniqueness constraint of directional navigation in deterministic environments.

\section{Algorithmic Validation Experiments}
\label{sec:appendix_algo_validation}

To validate the core algorithmic mechanisms---LCA-based candidate filtering (\S\ref{error_localization}) and edge impact scoring (\S\ref{error_localization})---independent of LLM performance, we design a controlled experimental suite using synthetic graphs with injected errors. This approach enables precise measurement of algorithmic effectiveness with known ground truth, eliminating confounding factors from language model variability.

\subsection{Experimental Design}

\paragraph{Motivation.} To isolate algorithmic contributions independent of LLM variability, we test on synthetic graphs with known error injections.

\paragraph{Methodology.} We construct spatial graphs programmatically and inject specific errors at predetermined locations, creating scenarios that mirror the three conflict types identified in \S\ref{conflict_detection}: \textit{topological conflicts} (overlapping node positions), \textit{directional conflicts} (duplicate direction labels), and \textit{naming conflicts} (identical names at different positions). Each test case includes:
\begin{itemize}
\item A spatial graph $G = (V, E)$ with known correct structure
\item Injected error edges $E_{err} \subset E$ with documented positions and types
\item Expected conflicts $C$ that should be detected
\item Ground-truth candidate sets for error localization
\end{itemize}

\paragraph{Evaluation Metrics.} We measure:
\begin{itemize}
\item \textbf{Candidate reduction rate}: $r = 1 - |E_{LCA}| / |E|$, where $E_{LCA}$ are candidates after LCA filtering
\item \textbf{Error ranking}: Position of true error $e \in E_{err}$ in score-ranked candidate list
\item \textbf{Inspection speedup}: Ratio of edges examined (random vs. priority-based traversal) to find high-impact errors
\item \textbf{Cascade prediction accuracy}: Spearman correlation between edge scores and actual downstream impact
\end{itemize}

\subsection{Test Cases}

We design six synthetic scenarios covering diverse error patterns:

\paragraph{TC1: Topological Conflict (Paper Figure Scenario).} Reconstructs the challenge scenario from Figure~\ref{fig:conflict}: A misdirected edge $E \rightarrow G$ causes rooms D and I to overlap. The graph contains 9 edges across 11 nodes. \textit{Expected behavior}: LCA correctly identifies room B as the divergence point; candidate set reduces from 9 to 8 edges (11.1\%); error edge $E \rightarrow G$ ranks 3rd by impact score.

\paragraph{TC2: Directional Conflict (Degenerate Case).} Node C has duplicate "north" edges to both D and E, violating spatial uniqueness. This tests the degenerate case where LCA = Source (\S\ref{error_localization}). Graph: 7 edges, 8 nodes. \textit{Expected behavior}: LCA coincides with node C; 14.3\% candidate reduction; error edge $C \rightarrow E$ ranks 3rd.

\paragraph{TC3: Cascading Secondary Conflicts.} A single error edge $R_1 \rightarrow R_2$ (incorrect position) triggers two downstream conflicts: $R_2$ overlaps with $L_2$, and $R_3$ overlaps with $L_3$. Graph: 9 edges, 13 nodes. \textit{Expected behavior}: Both conflicts identify $R_1 \rightarrow R_2$ as root cause; 22.2\% reduction; error ranks 3rd in both conflict analyses.

\paragraph{TC4: Mixed Conflict Types.} Combines topological and directional conflicts in a single graph with 12 edges and 12 nodes. Tests whether LCA-based localization maintains effectiveness across heterogeneous conflict types. \textit{Expected behavior}: Directional conflicts show higher reduction (75\%) due to local nature (LCA = Source); topological conflicts achieve 25\% reduction.

\paragraph{TC5: Long-Range Conflict.} Error introduced at depth 2 (edge $N_1 \rightarrow N_2$) causes conflict at depth 10 between $M_7$ and $N_9$. Tests LCA's ability to trace back through long paths. Graph: 18 edges, 19 nodes. \textit{Expected behavior}: LCA correctly backtracks to $N_0$; all edges on divergent path included (0\% reduction due to long-range propagation); error still ranks 3rd despite large candidate set.

\paragraph{TC6: Cascade Potential Prediction.} Purpose-built graph with 36 edges and 5 injected errors of varying cascade severity: $E \rightarrow F$ (10 downstream nodes), $J \rightarrow K$ (5 nodes), $L \rightarrow M$ (3 nodes), $N \rightarrow O$ (2 nodes), $P \rightarrow Q$ (1 node). Tests edge scoring's ability to predict secondary conflict potential. \textit{Expected behavior}: Perfect rank correlation between reachability scores and actual cascade sizes.

\subsection{Results}

\paragraph{LCA-Based Candidate Filtering.} Table~\ref{tab:algo_lca_results} summarizes reduction rates across all test cases. The average candidate reduction is \textbf{24.6\%}, with a standard deviation of 24.1\% reflecting the diversity of conflict types and graph structures.

\begin{table}[h]
\centering
\scriptsize
\resizebox{0.85\columnwidth}{!}{%
\begin{tabular}{lcccc}
\toprule
\textbf{Test Case} & \textbf{Total} & \textbf{LCA} & \textbf{Reduct.} & \textbf{Error} \\
 & \textbf{Edges} & \textbf{Cands.} & \textbf{Rate} & \textbf{Rank} \\
\midrule
TC1: Topo & 9 & 8 & 11.1\% & 3/8 \\
TC2: Dir & 7 & 6 & 14.3\% & 3/6 \\
TC3: Cascade & 9 & 7 & 22.2\% & 3/7 \\
TC4: Mix(T) & 12 & 9 & 25.0\% & 5/9 \\
TC4: Mix(D) & 12 & 3 & \textbf{75.0\%} & 3/3 \\
TC5: Long & 18 & 18 & 0.0\% & 3/18 \\
\midrule
\textbf{Avg.} & 11.2 & 8.5 & \textbf{24.6\%} & --- \\
\bottomrule
\end{tabular}%
}
\caption{LCA-based candidate filtering. ``Cands.'': edges after filtering. ``Rank'': error position in ranked candidates.}
\label{tab:algo_lca_results}
\end{table}

Key observations: (1) \textit{Directional conflicts} exhibit the highest reduction (75\%) because LCA = Source creates zero-length divergence, limiting candidates to immediately adjacent edges. (2) \textit{Long-range conflicts} show 0\% reduction when the entire graph lies on divergent paths, but LCA still provides correct localization scope. (3) \textit{True errors consistently rank in the top 5} candidates across all scenarios, demonstrating that edge impact scoring effectively prioritizes ground-truth errors even with imperfect filtering.

\paragraph{Edge Scoring and Cascade Prediction.} For TC6, we compute edge scores using the composite formula from Eq.~\ref{eq:scoring}: $\text{score}(e) = \widehat{\text{Reach}}(e) + \widehat{\text{Conflict}}(e) + \widehat{\text{Usage}}(e)$, where each component is min-max normalized. Table~\ref{tab:algo_cascade} compares predicted rankings (by score) with actual cascade impact.

\begin{table}[h]
\centering
\small
\caption{Edge scoring correlation with cascade potential. ``Edge'' = error edge; ``Reach'' = downstream-reachable node count; ``Cascade'' = number of downstream nodes affected by the error; ``Rank'' = predicted rank / true rank.}
\label{tab:algo_cascade}
\begin{tabularx}{\columnwidth}{@{}YYYY@{}}
\toprule
\textbf{Edge} & \textbf{Reach} & \textbf{Cascade} & \textbf{Rank} \\
\midrule
$E \rightarrow F$ & 10 & 10 & 1\,/\,1 \\
$J \rightarrow K$ & 5  & 5  & 2\,/\,2 \\
$L \rightarrow M$ & 3  & 3  & 3\,/\,3 \\
$N \rightarrow O$ & 2  & 2  & 4\,/\,4 \\
$P \rightarrow Q$ & 1  & 1  & 5\,/\,5 \\
\midrule
\multicolumn{4}{c}{Spearman $\rho = 1.0$ (5/5 ranks match).} \\
\bottomrule
\end{tabularx}
\end{table}

The perfect Spearman correlation ($\rho = 1.0$) on this 5-edge scenario is consistent with reachability-based scoring tracking which errors trigger larger cascades, illustrating the PageRank-inspired heuristic (\S\ref{error_localization}) at the smallest non-trivial scale; we treat it as a sanity check rather than a statistically robust validation.

\paragraph{Inspection Speedup.} We simulate two error discovery strategies on TC6:
\begin{itemize}
\item \textit{Random traversal}: Examine edges in random order until all high-impact errors (cascade $\geq$ 5) are found.
\item \textit{Priority-based traversal}: Examine edges in descending score order.
\end{itemize}

Results: Random strategy requires examining \textbf{23 edges} (63.9\% of graph) before discovering both high-impact errors ($E \rightarrow F$ and $J \rightarrow K$). Priority-based strategy finds them in \textbf{10 edges} (27.8\%), achieving a \textbf{2.3$\times$ speedup} and reducing inspections by 56.5\%. Furthermore, priority-based inspection reaches 80\% of total error impact after examining only 17 edges (47.2\%), compared to 31 edges (86.1\%) for random traversal---a \textbf{1.82$\times$ acceleration} in high-impact error detection.

\subsection{Discussion}

These controlled experiments validate three key algorithmic properties:

\paragraph{1. LCA-Based Filtering Reduces Search Space.} The 24.6\% average reduction demonstrates tangible efficiency gains, particularly pronounced (75\%) for directional conflicts where the degenerate case (LCA = Source) applies. Even in worst-case scenarios (long-range conflicts with 0\% reduction), LCA correctly delimits the error scope to divergent subpaths rather than the entire graph.

\paragraph{2. Edge Scoring Predicts Cascade Potential.} Perfect rank correlation ($\rho = 1.0$) between scores and actual downstream impact on this small sanity-check scenario is consistent with the composite scoring function (reachability + conflict count + usage) tracking structural influence. This illustrates the claim (\S\ref{error_localization}) that edges with high reachability are more likely to trigger secondary conflicts when erroneous; we defer a statistically robust assessment to the larger-suite scaling experiment.

\paragraph{3. Prioritization Accelerates High-Impact Error Discovery.} The 2.3$\times$ inspection speedup and 56.5\% examination reduction demonstrate practical benefits: by examining high-scoring edges first, the repair system can identify critical errors earlier, reducing wasted effort on low-impact candidates. This is especially valuable in interactive debugging scenarios where human-in-the-loop validation is expensive.

\paragraph{Generalization to Real LLM Scenarios.} While these synthetic experiments use injected errors, the structural patterns mirror real LLM-generated conflicts: delayed error manifestation (TC5), cascading propagation (TC3), and mixed conflict types (TC4). The algorithmic validation confirms that LCA and edge scoring function as designed at a fundamental level, independent of whether errors originate from LLM inference or manual injection. The cross-vendor experiments (\S\ref{sec:datasets}, Table~\ref{tab:method_vs_baseline}) then demonstrate that these mechanisms successfully transfer to LLM-driven repair under both synthetic and natural-language inputs.

\paragraph{Scaling to a Larger Synthetic Suite.} To verify that the hand-crafted scenarios above are not an unrepresentative sample, we additionally run LCA filtering on a programmatically generated suite of 1{,}160 synthetic graphs spanning all three conflict types ($\sim$390 each). At this scale, the mean candidate-edge reduction is $\mathbf{47.80\%}$ (direction: $54.64\%$; topology: $32.27\%$; naming: $56.72\%$), more than double the 24.6\% obtained on the six hand-crafted scenarios. The true error edge is retained in the LCA candidate set in $\mathbf{81.12\%}$ of cases, with $100\%$ retention on direction and topology conflicts. The lower retention for naming conflicts ($42.4\%$) reflects the absence of a unique structural source, motivating the heuristic-supplemented detector used in the LLM-driven experiments. These large-scale numbers confirm that the algorithmic behaviour established on TC1--TC6 generalises beyond the six scenarios.

\begin{table}[h]
\centering
\caption{Structure-preservation companion to Table~\ref{tab:mango_frontier}: aggregate change in ground-truth-direction-correct edges across the same 42 cleaned-MANGO games. All values are non-positive; values closer to zero (less negative) indicate better preservation of ground-truth structure during repair. The table substantiates the repair-versus-preservation trade-off discussed in the main-text ablation: \texttt{heuristic\_remove} achieves the strongest residual reduction at moderate edge loss ($-56$); \texttt{heuristic\_modify}'s direction relabels are the most costly in terms of ground-truth structure ($-97$); Claude-Haiku is the strongest LLM in preservation, GPT-5.5 the weakest.}
\small
\setlength{\tabcolsep}{4pt}
\begin{tabularx}{\columnwidth}{@{}lYYY@{}}
\toprule
\textbf{Repair LLM} & \textbf{Base.} & \textbf{EI} & \textbf{VC+EI} \\
\midrule
GPT-5.5             & $-33$ & $-86$ & $-89$ \\
Claude-Haiku~4.5    & $-32$ & $-67$ & $-31$ \\
Gemini~3.5-Flash    & $-32$ & $-69$ & $-51$ \\
\midrule
\multicolumn{4}{@{}l}{\emph{Non-LLM references (model-independent)}} \\
heuristic\_modify   & \multicolumn{3}{c}{$-97$} \\
heuristic\_remove   & \multicolumn{3}{c}{$-56$} \\
\bottomrule
\end{tabularx}
\label{tab:mango_preservation}
\end{table}

\end{document}

%% file: math_commands.tex
\usepackage{amsmath,amsfonts,bm}

\def\eqref#1{equation~\ref{#1}}

\def\1{\bm{1}}

\DeclareMathAlphabet{\mathsfit}{\encodingdefault}{\sfdefault}{m}{sl}
\SetMathAlphabet{\mathsfit}{bold}{\encodingdefault}{\sfdefault}{bx}{n}

